\documentclass[letterpaper, 10 pt, conference]{ieeeconf}  

\IEEEoverridecommandlockouts                              

\usepackage{array}
\usepackage{booktabs}
\usepackage{multirow}
\usepackage{xcolor}
\usepackage{makecell}
\usepackage{caption}
\usepackage{amsmath}
\usepackage{amssymb}
\usepackage{graphicx}
\usepackage{url}

\definecolor{ForestGreen}{RGB}{34,139,34}

\title{\LARGE \bf
PharmaShip: An Entity-Centric, Reading-Order–Supervised Benchmark for Chinese Pharmaceutical Shipping Documents
}

\author{Tingwei Xie, Tianyi Zhou, Yonghong Song$^{*}$ \\
  School of Software Engineering, Xi'an Jiaotong University, Xi'an, China
}

\begin{document}

\maketitle
\thispagestyle{empty}
\pagestyle{empty}

\begin{abstract}

We present PharmaShip, a real-world Chinese dataset of scanned pharmaceutical shipping documents designed to stress-test pre-trained text–layout models under noisy OCR and heterogeneous templates. PharmaShip covers three complementary tasks—sequence entity recognition (SER), relation extraction (RE), and reading order prediction (ROP)—and adopts an entity-centric evaluation protocol to minimize confounds across architectures. We benchmark five representative baselines spanning pixel-aware and geometry-aware families (LiLT, LayoutLMv3-base, GeoLayoutLM and their available RORE-enhanced variants), and standardize preprocessing, splits, and optimization. Experiments show that pixels and explicit geometry provide complementary inductive biases, yet neither alone is sufficient: injecting reading-order–oriented regularization consistently improves SER and EL and yields the most robust configuration, while longer positional coverage stabilizes late-page predictions and reduces truncation artifacts. ROP is accurate at the word level but challenging at the segment level, reflecting boundary ambiguity and long-range crossings. PharmaShip thus establishes a controlled, reproducible benchmark for safety-critical document understanding in the pharmaceutical domain and highlights sequence-aware constraints as a transferable bias for structure modeling. We release the dataset at \url{https://github.com/KevinYuLei/PharmaShip}.

\end{abstract}

\section{INTRODUCTION}

The task of visually-rich document understanding (VrDU) has garnered increasing attention with the growing industrial need for automating information extraction from complex document layouts\cite{10.1007/978-981-99-7894-6_10}. In domains such as healthcare, finance, and logistics, scanned documents with irregular structures and domain-specific semantics present significant challenges to pre-trained text-and-layout models (PTLMs). While prior efforts have led to the creation of datasets such as FUNSD\cite{jaume2019}, CORD\cite{park2019cord}, and SROIE\cite{8977955}, these benchmarks primarily focus on general administrative forms or receipts. Early related end-to-end models\cite{Appalaraju_2021_ICCV}, pre-trained methods\cite{hong_bros_2022}, multimodal methods\cite{Xu2020LayoutLMPO, li2021structext}, and layout awareness distribution models\cite{DBLP:conf/cvpr/GuMWLWG022} in the field of VrDU thus emerged.

In the pharmaceutical industry, delivery documents, including shipment notes, outbound order slips, and accompanying logistics receipts, play a critical role in ensuring accurate medication tracking, compliance with regulatory standards, and transparent inventory flow\cite{MANN2023108446}. These documents often contain crucial structured fields such as drug names, dosages, quantities, batch numbers, expiry dates, and logistic codes\cite{103e023c5c55422ba35b1074fc230132}. Extracting and linking information from complex pharmaceutical documents benefits digital archiving, auditing, and advances in visually-rich document understanding for pharmacovigilance and logistics.\cite{li2020docbank} However, to the best of our knowledge, there is no publicly available dataset tailored to this document class.

To address this gap, we introduce PharmaShip, a scan-based dataset of pharmaceutical delivery documents, specifically curated for training and evaluating models on Semantic Entity Recognition (SER), Entity Linking (EL), and Reading Order Prediction (ROP) tasks. The dataset is inspired by and extends the design principles of EC-FUNSD\cite{Zhang2024UnveilingTD}. Following the entity-centric annotation paradigm, PharmaShip provides both segment and word-level annotations for semantic entities and their relations, while also incorporating a directed acyclic reading order graph to capture the layout-induced reading strategies\cite{6628706}, similar to the Immediate Succession During Reading (ISDR) formulation introduced in ROOR\cite{zhang2024modeling}.

Despite steady progress in visually rich document understanding, benchmarks curated for generic forms and consumer receipts fall short for pharmaceutical shipping. They cover few medicine-specific fields, exhibit limited layout variability, and rarely encode reading flow—signals needed to disambiguate multi-column tables and dense text. We therefore introduce PharmaShip, a real-world Chinese corpus that targets these gaps with entity-centric semantics, fine-grained relations, and long-form layouts. In summary, our contributions are threefold:

1. We construct PharmaShip, the first benchmark dataset of pharmaceutical delivery documents;

2. We annotate semantic entities, relation triples, and reading order graphs at the segment and word level, enabling fine-grained layout-aware training on SER, EL, and ROP tasks;

3. We benchmark multiple state-of-the-art models, demonstrating the dataset's utility in advancing VrDU in real-world healthcare logistics.

\section{RELATED WORK}

\subsection{Visually-rich Document Understanding Datasets}

Classic VrDU datasets such as FUNSD, CORD, and SROIE primarily comprise administrative forms or receipts with fixed templates and block-level annotations, which induce segment–entity coupling and favor proximity-based heuristics; reading-order supervision is typically absent. Entity-centric variants (e.g., EC-FUNSD) alleviate part of this bias but still lack long-form documents and domain-specific semantics typical of pharmaceutical logistics. Against this backdrop, PharmaShip serves as a complementary benchmark that emphasizes sequence awareness and robustness to layout drift without reiterating generic receipt assumptions\cite{peng-etal-2022-ernie}.

PharmaShip focuses on scanned pharmaceutical delivery documents characterized by complex layouts, multi-line field groupings, and information distributed across tables and headers. Unlike previous datasets, it provides fine-grained semantic entity labels, directed inter-entity relations, and document-level reading order graphs in a unified annotation scheme. This design makes PharmaShip a more realistic and comprehensive benchmark for evaluating models on SER, EL, and ROP tasks in layout-intensive, industry-grade document scenarios.

\subsection{Reading Order Prediction}

Reading Order Prediction (ROP) aims to identify the intended reading flow of textual segments in complex layouts. While traditional approaches represent reading order as a linear sequence\cite{quiros_reading_2022}, recent work reformulates it as a graph-based relation extraction task to better reflect non-linear and multi-directional reading behavior\cite{zhang-etal-2023-reading}. The ISDR (Immediate Succession During Reading) formulation models this reading flow as a directed acyclic graph, which has proven effective in improving downstream tasks such as entity recognition and relation extraction.

PharmaShip adopts this improved formulation by explicitly annotating reading order as graph edges between visual segments. In dense tabular layouts—such as those found in pharmaceutical delivery notes—key fields like drug name, specification, quantity, and unit price are visually adjacent but semantically ordered. By using ISDR-style links, PharmaShip provides accurate reading order supervision, enhancing the structural interpretability of documents and supporting more robust layout-aware models in real-world scenarios.

\subsection{Entity-Centric Evaluation}
Many existing VrDU datasets rely on layout-driven or block-level annotations, where semantic entities are implicitly tied to visual regions. While sufficient for basic structure parsing, this strategy introduces ambiguity when visually adjacent elements belong to different semantic categories. Such layout-centric designs can mislead models into learning superficial spatial patterns rather than true semantic distinctions, limiting their generalization to complex or domain-specific scenarios.

PharmaShip adopts an entity-centric annotation approach that explicitly separates semantic labeling from layout structure. Entities are annotated at the segment level with precise semantic types, regardless of their visual proximity. This design ensures that conceptually distinct elements—such as drug names, batch numbers, and expiration dates—are consistently and accurately labeled. By decoupling semantics from layout, PharmaShip reduces annotation bias, enables finer-grained model supervision, and better aligns with real-world information extraction needs.

\section{DATASET CONSTRUCTION}
\subsection{Data Statistics}
The PharmaShip dataset is constructed from real-world pharmaceutical logistics documents collected from multiple enterprises. It includes scanned delivery notes, accompanying shipment forms, and outbound sales slips—document types commonly used in China’s pharmaceutical circulation and inventory management systems. These documents predominantly feature dense tabular layouts, yet crucial information—such as recipient details, document numbers, and dispatch dates—often appears outside the main table regions, typically in the header or footer. This layout heterogeneity poses additional challenges for semantic entity recognition and reading order prediction tasks.

The dataset consists of 161 annotated scanned documents, each provided in JSON format with structured annotations for semantic entities, entity-level relations, and segment-level reading order links. Compared with datasets like ROOR (which is an extension of EC-FUNSD on reading order), PharmaShip provides more domain-specific complexity. Detailed statistics, including the number of entities, relations, and segment length distributions, are summarized in Table~\ref{tab:stats}.

\begin{table*}[ht]
\caption{Statistics of PharmaShip, ROOR, FUNSD, CORD, and SROIE, including words, segments, entities, relation pairs, and the presence/strength of reading-order supervision.}
\label{tab:stats}
\scriptsize

\makebox[\textwidth][c]{
\begin{tabular}{>{\centering\arraybackslash}m{2cm}
                >{\centering\arraybackslash}m{2cm}
                >{\centering\arraybackslash}m{2cm}
                >{\centering\arraybackslash}m{2cm}
                >{\centering\arraybackslash}m{2cm}
                >{\centering\arraybackslash}m{2cm}}
\toprule
\textbf{Dataset} & 
\shortstack{\textbf{\#Segments}} & 
\shortstack{\textbf{\#Segs.}\\\textbf{Words}} & 
\shortstack{\textbf{\#Segs. per}\\\textbf{Sample}} & 
\shortstack{\textbf{\#Segs. Words}\\\textbf{per Sample}} & 
\shortstack{\textbf{\#Avg. Len.}\\\textbf{of Segment}} \\
\midrule
FUNSD\cite{jaume2019}           & 9,743  & 31,485 & 48.96 & 158.22 & 3.23 \\
CORD\cite{park2019cord}         & 12,499 & 21,872 & 13.63 & 23.85  & 1.75 \\
SROIE\cite{8977955}             & 33,626 & 72,390 & 53.72 & 115.64 & 2.15 \\
ROOR\cite{zhang2024modeling}    & 10,662 & 31,297 & 53.57 & 157.27 & 2.93 \\
PharmaShip                      & 11,295 & 66,961 & 70.16 & 415.91 & 5.93 \\
\end{tabular}
}

\vspace{0em}

\makebox[\textwidth][c]{
\begin{tabular}{>{\centering\arraybackslash}m{2cm}
                >{\centering\arraybackslash}m{2cm}
                >{\centering\arraybackslash}m{2cm}
                >{\centering\arraybackslash}m{2cm}
                >{\centering\arraybackslash}m{2cm}
                >{\centering\arraybackslash}m{2cm}}
\toprule
\textbf{Dataset} & 
\shortstack{\textbf{\#Entities}} & 
\shortstack{\textbf{\#Ents.}\\\textbf{Words}} & 
\shortstack{\textbf{\#Ents. per}\\\textbf{Sample}} & 
\shortstack{\textbf{\#Ents. Words}\\\textbf{per Sample}} & 
\shortstack{\textbf{\#Avg. Len.}\\\textbf{of Entity}} \\
\midrule
FUNSD\cite{jaume2019}           & 9743   & 31,485 & 48.96 & 158.22 & 3.23 \\
CORD\cite{park2019cord}         & 12,499 & 21,872 & 13.63 & 23.85  & 1.75 \\
SROIE\cite{8977955}             & 33,626 & 72,390 & 53.72 & 115.64 & 2.15 \\
ROOR\cite{zhang2024modeling}    & 8,398  & 24,873 & 42.20 & 124.99 & 2.96 \\
PharmaShip                      & 10,090 & 65,521 & 62.67 & 406.96 & 6.49 \\
\end{tabular}
}

\vspace{0em}

\makebox[\textwidth][c]{
\begin{tabular}{>{\centering\arraybackslash}m{2cm}
                >{\centering\arraybackslash}m{2cm}
                >{\centering\arraybackslash}m{2cm}
                >{\centering\arraybackslash}m{2cm}
                >{\centering\arraybackslash}m{2cm}
                >{\centering\arraybackslash}m{2cm}}
\toprule
\textbf{Dataset} & 
\shortstack{\textbf{\#QA}\\\textbf{Relations}} & 
\shortstack{\textbf{\#QA Rels.}\\\textbf{per Sample}} & \shortstack{\textbf{\#RO}\\\textbf{Relations}} & 
\shortstack{\textbf{\#RO Rels.}\\\textbf{per Sample}} & 
{} \\
\midrule
FUNSD\cite{jaume2019}           & 10,624 & 53.39 & -- & --  & {} \\
ROOR\cite{zhang2024modeling}    & 5,200  & 26.13 & 10,966 & 55.11  & {} \\
PharmaShip                      & 5,005  & 31.09 & 17,419 & 108.19 & {} \\
\bottomrule
\end{tabular}
}
\end{table*}

\begin{figure}
    \centering
    \includegraphics[width=1.0\linewidth]{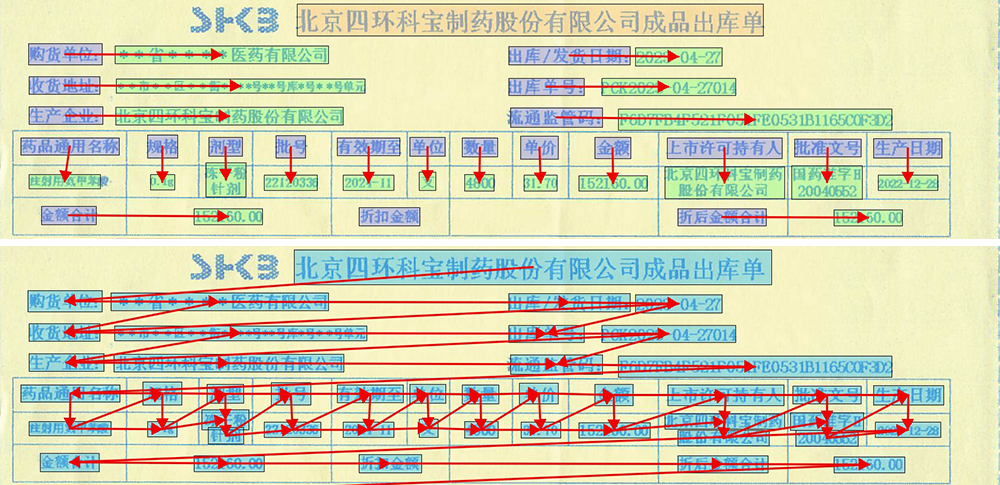}
    \caption{An example page from the PharmaShip dataset. Top: annotations for SER and EL, with \textit{headers} in orange, \textit{questions} in blue, and \textit{answers} in green. Bottom: ROP represented as Immediate Successor During Reading (ISDR) edges forming a directed acyclic graph over all segment-level semantic entities.}
    \label{fig:example}
\end{figure}

Table \ref{tab:stats} contrasts PharmaShip with ROOR and classic VrDU corpora (FUNSD, CORD, SROIE). With a similar sample size to ROOR, PharmaShip contains more segments and over twice the segment words, alongside higher entity and relation density and nearly double the average reading-order links ($\approx$108 vs. $\approx$55). Relative to FUNSD/CORD/SROIE, PharmaShip further stands out for long-form layouts and explicit reading-order supervision. Figure \ref{fig:example} illustrates a labeled PharmaShip page.

\subsection{Task Formulation}

PharmaShip supports three major tasks: Semantic Entity Recognition (SER), Relation Extraction (RE), and Reading Order Prediction (ROP). Each task corresponds to a distinct part of the dataset structure.

\textbf{SER} is defined as a token-level classification task. Formally, given a document layout with $N$ visual segments:
\begin{equation}
D = \left\{ (w_i, b_i) \right\}_{i=1}^{N}
\end{equation}

where $w_i$ denotes the textual content of the (i)-th segment and $b_i \in \mathbb{R}^4$ is its bounding box, the goal of SER is to learn a label function:
\begin{equation}
f_{\text{SER}}: w_i \mapsto l_i \in \mathcal{L}
\end{equation}

where $\mathcal{L}$ is the set of entity types (e.g., "header", "question", "answer", "other"). SER evaluates recognizing semantic entities and grounding each entity to its corresponding textual segment in the layout, decoupling entities from block-level annotations.

\textbf{EL} evaluates predicting valid links between recognized entities and assigning the correct relation type for each subject–object pair. Given a set of recognized entities $E=\{e_i\}_{i=1}^M$, the task seeks to predict relation triplets:
\begin{equation}
f_{\text{RE}}: (e_i, e_j) \mapsto r_{ij} \in \mathcal{R}
\end{equation}

where $\mathcal{R}$ denotes the "question"-"answer" or other valid relation pairs. 

\textbf{ROP} evaluates recovering a coherent reading sequence over segments that reflects the human reading flow across the document. ROP is modeled as a relation extraction problem over layout segments. Following Zhang et al. (2024), in the document defined as $\mathcal{D}=\{s_i\}_{i=1}^N$, the task predicts a set of ISDR links:
\begin{equation}
    f_{ROP}: (s_i, s_j) \mapsto p_{ij} \in \mathcal{P}
\end{equation}

where $\mathcal{P}$ denotes a set of immediate succession links. This forms a directed acyclic graph over segment indices. 

\section{Experiments}
\subsection{Baseline Methods}
We evaluate PharmaShip with five representative baselines chosen to reflect the entity-centric evaluation philosophy advocated for visually rich documents while controlling confounds across architectures. Concretely, we include LiLT[InfoXLM]\cite{wang-etal-2022-lilt}, LayoutLMv3-base-chinese\cite{huang2022layoutlmv3}, GeoLayoutLM\cite{cvpr2023geolayoutlm}, and their reading-order counterparts, namely RORE-LayoutLMv3, and RORE-GeoLayoutLM, using official implementations and Chinese checkpoints where available. LiLT serves as a multilingual text encoder with layout cues injected via box/position interactions, enabling us to probe a pixel-free layout induction regime. LayoutLMv3 jointly models tokens and rendered pixels, offering a pixel-aware baseline; RORE-LayoutLMv3 augments it with reading-order-relation enhanced (RORE) under the ROOR framework to discourage implausible token paths and spurious relations. GeoLayoutLM contributes explicit multi-level geometric encodings and relative spatial biases; its RORE variant unifies geometry with sequence-level guidance.

\subsection{Implementation Details}
We implement all models using the official releases of LiLT, LayoutLMv3-base-chinese, and GeoLayoutLM with their Chinese pre-trained checkpoints, and we keep the preprocessing pipeline identical across methods (same OCR tokens and bounding boxes; identical page renderings for pixel-aware models). Because the default maximum textual sequence length of 512 in these implementations is frequently exceeded by Chinese business documents, naïve truncation would remove late-page entities and long-range dependencies, degrading both SER and EL. To prevent this, we expand the positional embedding length from 512 to 2048 for all three backbones and fine-tune end-to-end, ensuring that no training or validation sample is truncated. Newly added positions are initialized following each repository’s default policy, and dynamic padding within mini-batches is used to control memory. This modification preserves complete context for long forms in PharmaShip while keeping the rest of the architecture unchanged, thereby isolating the effect of length coverage rather than capacity.

\begin{table*}[t]
\centering
\caption{Cross-dataset F1 performance of representative VrDU models on public benchmarks and PharmaShip. “Base” denotes parameter counts between 100–200M.}
\label{tab:x-dataset-results}
\scriptsize
\setlength{\tabcolsep}{5pt}               
\renewcommand{\arraystretch}{1.15}
\begin{tabular*}{0.8\textwidth}{@{\extracolsep{\fill}} l c cc c c cc cc @{}}
\toprule
\multirow{2}{*}{\textbf{\ Model}} & \multirow{2}{*}{\textbf{Params}} &
\multicolumn{2}{c}{\textbf{FUNSD}} & 
\multicolumn{1}{c}{\textbf{CORD}} &
\multicolumn{1}{c}{\textbf{SROIE}} &
\multicolumn{2}{c}{\textbf{ROOR}} &
\multicolumn{2}{c}{\textbf{PharmaShip}} \\
\cmidrule(lr){3-4}\cmidrule(lr){5-5}\cmidrule(lr){6-6}\cmidrule(lr){7-8}\cmidrule(lr){9-10}
 &  & SER & EL & SER & SER & SER & EL & SER & EL\  \\
\midrule
\ LiLT[InfoXLM]\cite{wang-etal-2022-lilt}          & base     & 85.74 & 74.07 & 95.77 & --    & --    & --    & 75.85 & 73.61\  \\
\ LayoutLMv3 \cite{huang2022layoutlmv3}            & 133M     & 90.85 & 69.80 & 95.91 & 94.80 & 82.30 & 67.47 & 83.49 & 77.25\  \\
\ RORE-LayoutLMv3\cite{zhang2024modeling}        & 133M+12  & 91.39 & 71.69 & 96.72 & 95.65 & 82.80 & 73.64 & 84.06 & 78.53\  \\
\ GeoLayoutLM\cite{cvpr2023geolayoutlm}            & 399M     & 91.10 & 88.06 & 98.11 & 96.62 & 83.62 & 86.18 & 80.69 & 75.52\  \\
\ RORE-GeoLayoutLM\cite{zhang2024modeling}       & 399M+24  & 91.84 & 88.46 & 98.52 & 96.97 & 84.34 & 87.42 & 81.88 & 76.56\  \\
\bottomrule
\end{tabular*}
\end{table*}

For training, we follow the initial settings of Huang et al.\cite{huang2022layoutlmv3} for LiLT and LayoutLMv3 on SER and EL, but we correct the class-imbalance issue observed in the public ROOR implementation that leads to overfitting toward the “no-relation” class. Specifically, we rebalance positive/negative pairs and apply class-weighted optimization to stabilize gradients. For GeoLayoutLM on EL, we adopt the schedule in Zhang et al.\cite{zhang2024modeling}, using AdamW with weight decay 1e$-$2, learning rate 1e$-$4, batch size 16, and 2\% linear warm-up, followed by a linear decay. All tasks are fine-tuned for up to 500 epochs with early stopping; training halts if no validation improvement is observed for 50 epochs, and we retain the checkpoints that achieve the best F1 on SER and EL. Unless a method requires otherwise, optimization hyperparameters and random seeds remain aligned to ensure comparability.

\subsection{Results}
Table \ref{tab:x-dataset-results} reports cross-dataset evaluations where we train and test identical model families on PharmaShip and conventional VrD benchmarks. While traditional receipts/forms yield higher absolute scores, the same architectures suffer clear degradations on PharmaShip—both SER and EL typically drop by 5–9 F1, despite preserving the ranking and steady gains from reading-order–enhanced variants. We attribute this to the entity-centric annotation principle we adopt—aligned with EC-FUNSD—which decouples segments from semantic entities and removes the block-level “false coupling” that encourages overfitting on benchmarks like FUNSD; once this bias is stripped, the fragility of layout heuristics becomes visible. In addition, PharmaShip contains fine-grained, long-range medical relations (e.g., lot–dosage–spec) whose resolution hinges on reliable reading order rather than local proximity. By reducing paragraph–entity coupling and stressing sequence awareness and robustness to layout drift, PharmaShip complements existing corpora while posing a harder, more diagnostic test bed that better matches real Chinese pharmaceutical shipping workflows.

\begin{table}[ht]
\caption{Performance comparison of different models on SER, EL, and ROP tasks of PharmaShip. Improvements (↑) denote F1 gains of RORE-enhanced variants.}
\label{tab:results}
\scriptsize
\setlength{\tabcolsep}{4pt}
\renewcommand{\arraystretch}{1.15}

\centering
\begin{tabular*}{\columnwidth}{@{\extracolsep{\fill}}@{} ll lccc @{}}
\toprule
\textbf{\ \ Task} & {} & \textbf{Model} & \textbf{Precision} & \textbf{Recall} & \textbf{F1} \\
\midrule
\multirow{4}{*}{\centering \ \ SER} 
  &  & LiLT[InfoXLM]\cite{wang-etal-2022-lilt}         & 73.15 & 78.77 & 75.85 \\
  &  & LayoutLMv3\cite{huang2022layoutlmv3}            & 81.06 & 86.08 & 83.49 \\
  &  & RORE-LayoutLMv3\cite{zhang2024modeling}       & 82.13 & 86.08 & \makecell[c]{84.06 \\ (\textcolor{ForestGreen}{↑0.57})} \\
  &  & GeoLayoutLM\cite{cvpr2023geolayoutlm}           & 77.71 & 83.91 & 80.69 \\
  &  & RORE-GeoLayoutLM\cite{zhang2024modeling}      & 78.81 & 85.21 & \makecell[c]{81.88 \\ (\textcolor{ForestGreen}{↑1.19})} \\
\midrule
\multirow{4}{*}{\centering \ \ EL} 
  &  & LiLT[InfoXLM]\cite{wang-etal-2022-lilt}         & 51.22 & 82.35 & 63.16 \\
  &  & LayoutLMv3\cite{huang2022layoutlmv3}            & 67.49 & 90.32 & 77.25 \\
  &  & RORE-LayoutLMv3\cite{zhang2024modeling}       & 68.81 & 91.45 & \makecell[c]{78.53 \\ (\textcolor{ForestGreen}{↑1.28})} \\
  &  & GeoLayoutLM\cite{cvpr2023geolayoutlm}           & 77.91 & 73.28 & 75.52 \\
  &  & RORE-GeoLayoutLM\cite{zhang2024modeling}      & 78.42 & 74.79 & \makecell[c]{76.56 \\ (\textcolor{ForestGreen}{↑1.04})} \\
\midrule
\multirow{2}{*}{\centering \ \ ROP} 
  & word-level     & LayoutLMv3\cite{huang2022layoutlmv3}        & 81.91 & 96.20 & 88.48 \\
\cmidrule(l){2-6}  
  & segment-level  & LayoutLMv3\cite{huang2022layoutlmv3}        & 62.64 & 81.51 & 70.84 \\
\bottomrule
\end{tabular*}
\end{table}

The results of different methods on PharmaShip are shown in Table~\ref{tab:results} . We find that LayoutLMv3-base-chinese provides a strong pixel-aware baseline on PharmaShip for SER, and augmenting it with RORE produces the best SER performance. GeoLayoutLM starts lower, yet benefits more from RORE, narrowing the gap and indicating that sequence-level guidance complements explicit geometric biases. On EL, RORE-LayoutLMv3 achieves the highest F1 and both backbones exhibit consistent gains once reading order is injected, supporting the view that sequence-aware constraints mitigate spurious links induced by proximity-only cues. LiLT trails all layout–vision variants, underscoring the utility of pixel/geometry signals on Chinese shipping forms. For ROP, word-level ordering is largely reliable, while segment-level ordering remains challenging—an effect we attribute to boundary ambiguity and occasional long-range crossings in dense layouts.

Beyond absolute scores, the precision–recall profile clarifies the mechanism: geometry-only models tend to be precision-leaning, pixel-aware models recall-leaning, and RORE rebalances both by discouraging implausible token paths and enforcing coherent edges. Expanding positional capacity to 2048 prevents truncation on long forms, stabilizing performance and reducing late-page errors, which in turn further amplifies RORE’s benefits. With preprocessing, splits, and optimization aligned across methods, these trends remain robust. Taken together, our results echo recent entity-centric findings: naïve layout encodings can overfit dataset idiosyncrasies, whereas explicit modeling of reading order offers a transferable inductive bias for real-world pharmaceutical shipping documents.

\section{CONCLUSIONS}
We introduced PharmaShip, a real-world Chinese pharmaceutical shipping dataset, and provided a controlled evaluation of VrDU baselines. Our experiments reveal that pixel features and explicit geometry offer complementary inductive biases, yet neither alone is sufficient: injecting reading-order–oriented regularization systematically improves SER and EL and yields the most robust configuration, while expanding positional capacity to 2048 removes truncation artifacts that otherwise mask late-page errors. These findings echo entity-centric observations in recent work and highlight the value of sequence-aware constraints for reliable structure modeling under noisy OCR and heterogeneous templates. Remaining challenges include segment-level reading order and long-range dependencies across dense, multi-field pages. We release PharmaShip and our implementations to facilitate reproducible research and to catalyze progress on practical, safety-critical document understanding.

\addtolength{\textheight}{-9cm}   






\bibliographystyle{IEEEtran}
\bibliography{refs}


\end{document}